\documentclass{article}
\usepackage{iclr2027_conference,times}

\usepackage{amsmath,amsfonts,bm}

\def\eqref#1{equation~\ref{#1}}

\def\1{\bm{1}}

\DeclareMathAlphabet{\mathsfit}{\encodingdefault}{\sfdefault}{m}{sl}
\SetMathAlphabet{\mathsfit}{bold}{\encodingdefault}{\sfdefault}{bx}{n}

\usepackage{hyperref}
\usepackage{url}
\usepackage{booktabs}
\usepackage{array}   
\usepackage{graphicx}
\usepackage{float}
\usepackage[most]{tcolorbox}
\usepackage[T1]{fontenc}

\newcommand{\NumSamples}{2{,}197}
\newcommand{\NumFailures}{1{,}176}
\newcommand{\NumSuccesses}{1{,}021}
\newcommand{\FailPct}{54\%}
\newcommand{\NumSources}{fourteen}
\newcommand{\NumRealSources}{twelve}
\newcommand{\NumSimSources}{two}
\newcommand{\NumModels}{thirteen}

\newcommand{\BestScore}{0.77}
\newcommand{\BestModel}{Gemini 3 Flash}

\newcommand{\SimRealRho}{0.95}
\newcommand{\SimMinusReal}{1.3}

\newcommand{\WorldGymRTOne}{0.89}

\newcommand{\RoboFacSelfReport}{80.6\%}

\newcommand{\CropGeminiGemini}{+2.3}

\iclrfinalcopy

\begin{document}

\begin{tcolorbox}[
  enhanced,
  colback=black!4,
  frame hidden,
  arc=8pt,
  boxrule=0pt,
  left=18pt, right=18pt, top=16pt, bottom=16pt,
  width=\textwidth
]
\begin{center}
{\LARGE\sc\bf FailBench: How Reliable are VLMs at Judging Robot Task Success?\par}
\vskip 0.2in
\begin{tabular}[t]{c}
  Zaruhi Navasardyan \quad Tatul Danielyan \quad Hrant Davtyan \\[4pt]
  Metric AI Lab \\
  \texttt{\{zaruhi,tatul,hrant\}@metricailab.com}
\end{tabular}
\end{center}
\vskip 0.25in
\noindent
Vision-language models (VLMs) are increasingly used to determine whether a robot manipulation attempt has succeeded. These judgments can serve as reinforcement-learning rewards, training-data filters, policy-ranking signals, or triggers for retry, making reliable failure detection critical for robot learning and evaluation. However, existing benchmarks provide limited evidence of cross-domain generalization. We introduce FailBench, a benchmark for robot failure detection comprising 2,197 manipulation attempts from 14 public sources, including 12 real-world and 2 simulated datasets, using the original outcome labels provided by their respective sources. Six of the real-world sources were originally collected for policy evaluation, reward modeling, or general data collection rather than failure detection, and 75\% of failures occur naturally rather than being deliberately constructed. Evaluating 13 VLM-based detectors, we find that the best model achieves only 0.77 mean balanced accuracy, indicating substantial room for improvement. Models fine-tuned specifically for robot failure detection consistently underperform general-purpose VLMs, with the exception of the smallest general-purpose model, and also underperform their corresponding pretrained models. We further find that performance depends more strongly on the visual evidence required to determine the outcome than on the robot or task domain: detection approaches saturation when success is determined by observable object motion, but approaches chance when success depends on establishing contact, with no model exceeding 0.60 balanced accuracy on contact-intensive assembly tasks. Analysis of model reasoning reveals a systematic bias toward predicting success under ambiguous evidence. Increasing reasoning effort does not alleviate this bias, as incorrect predictions tend to receive longer reasoning traces. Model-level interventions provide little improvement, while an input-level intervention is effective: spatially localizing the outcome-relevant region and cropping the input improves the strongest detector by 2.4 percentage points without additional training. We release FailBench and the accompanying evaluation harness to facilitate reproducible evaluation of robot failure detection.
\\
\\
\textbf{Project Home:} \href{https://metric-ai-lab.github.io/failbench/}{https://metric-ai-lab.github.io/failbench/}
\end{tcolorbox}

\section{Introduction}

A single robot-learning experiment can produce hundreds or thousands of rollouts. To
evaluate a policy, filter the resulting data, or learn from it, the system must determine
which attempts succeeded. Reviewing every rollout manually is costly, so recent systems
increasingly assign this decision to a vision-language model given the task instruction and
a recording. The model's answer is then used as an evaluation label, a data filter, a
reward, or a recovery signal---typically as though it were ground truth
\citep{vlarl2025,soar2024,autoeval2025,reflect2023}. What remains unclear is whether these
judgments are reliable beyond the data on which the detector was developed.

Failure detectors take two broad forms. Policy-dependent monitors infer failure from a
policy's internal representations, action statistics, or predicted trajectories, but are
therefore tied to that policy and its architecture
\citep{sentinel2024,fiper2025,foresight2026,vlafail2026}. Policy-agnostic detectors instead
judge a completed attempt from its recording and instruction. Unlike a policy-dependent
monitor, the same detector can in principle be applied across policies and robot platforms.
This generality is useful, but it also makes transfer central: an incorrect verdict can
become a noisy reward, a misleading evaluation, or a failed trajectory retained as training
data.

Recent work increasingly asks VLMs for more than a binary verdict: an explanation
\citep{reflect2023,aha2024,armor2026}, a failure category and a corrective action
\citep{guardian2025,robofac2025,vifailback2025,failsafe2025,cfrecovery2026,promptopt2024}, a
subgoal-level or dense progress signal
\citep{stepeval2025,prmjudge2026,primo2026,robometer2026,progressreward2026}, or a graded
score rather than a pass/fail label
\citep{evalactions2026,roboeval2025,armnetbench2026,roboreward2026}. These richer outputs
all rely on the same basic capability: determining whether the attempted task succeeded.
That capability has not been evaluated systematically across data sources; where it has
been measured, evaluation has largely remained within the source used to develop the
detector.

Existing failure benchmarks do not resolve this gap. Although several are larger than
FailBench \citep{guardian2025,robofac2025,vifailback2025,aha2024}, each is drawn from a
single collection effort. Most also create failures deliberately, for example by
perturbing a successful trajectory or pairing an unchanged recording with a different
instruction. Such constructions are useful for controlled evaluation, but they can make
failure detection partly a matter of recognizing one data-generation procedure. The
failures encountered in ordinary robot operation are often less conspicuous: a peg that
stops just short of insertion, a gripper that appears closed but never establishes contact,
or a part released slightly too early.

We introduce FailBench, a benchmark of \NumSamples{} robot manipulation attempts collected
from \NumSources{} public sources, including \NumRealSources{} real-world and
\NumSimSources{} simulated sources. Each attempt is represented by an instruction, a visual
observation, and a binary outcome. Six of the real-world subsets come from existing failure
benchmarks. The remaining six come from datasets originally collected for policy
evaluation, reward-model evaluation, or general-purpose robot learning rather than failure
detection \citep{rh20t2024,reassemble2025,roboarena2025,armnetbench2026,robometer2026,
roboreward2026}. We use the labels assigned by the original data providers. The simulated
subsets contain rollouts of real-robot policies and are included because simulation remains
a primary setting for training and evaluating failure detectors.

FailBench focuses on \emph{execution failures}: cases in which an attempted action does not produce the intended outcome and the result can be judged from the recording. It does not cover \emph{planning failures}, in which the intended action is already inappropriate before execution, nor does it evaluate intermediate task progress \citep{progressreward2026}. This narrow scope gives us a common and directly comparable question: given the instruction and the visual recorded attempt, did the task succeed?

The results show that current detectors remain unreliable outside their original evaluation settings. The best of the \NumModels{} models reaches only \BestScore{} mean balanced
accuracy. Every detector fine-tuned specifically for robot failure detection scores below every general-purpose model except the smallest, and all 5 specialists that allow a direct comparison perform worse than their own base models. Performance depends strongly on the evidence required to determine the outcome. Detectors are reliable when failure can be inferred from which object moved, but approach chance when success depends on whether physical contact was established; no model exceeds 0.60 balanced accuracy on contact-rich assembly. Guided by this diagnosis, we test an evidence-localization pipeline that crops the visual input to the region that determines the outcome. This raises the strongest detector's mean balanced accuracy
by \CropGeminiGemini{} points without retraining it.

\paragraph{Contributions.}
\begin{itemize}
\item FailBench: \NumSamples{} robot execution attempts from \NumSources{} public sources in a common schema, spanning policy rollouts, human teleoperation, and constructed failures with both single and multi-view observations.
\item A cross-source evaluation of \NumModels{} failure detectors, using each model's intended prompt and inference settings.
\item An analysis of where and why these detectors fail, connecting their errors to the
visual evidence required to distinguish coarse object motion from physical contact.
\item An evidence-localization pipeline that improves failure detection by directing the
model to the region of the recording that determines the outcome.
\end{itemize}

\section{Related Work}
\label{sec:related}

Current research already uses vision-language models as judges of whether a robot succeeded. It is used as the reward that trains a policy \citep{vlarl2025,grape2024}, the training data filter \citep{soar2024}, the evaluator
that ranks policies \citep{autoeval2025,stepeval2025,worldgym2025}, and the monitor that
decides whether to retry \citep{reflect2023,codeasmonitor2024,promptopt2024}. In every one of those jobs, the answer is used as ground truth, and the quality of the judge itself is rarely evaluated extensively before it is integrated into the pipeline. Nonetheless, AutoEval correlates its whole pipeline against human evaluation without isolating the classifier \citep{autoeval2025}; WorldGym reports \WorldGymRTOne{} balanced accuracy for GPT-4o on RT-1 labels, then applies that judge to generated rollouts \citep{worldgym2025}. Text judges have the same gap, used widely and checked rarely \citep{llmjudges2024,assessingjudges2025}, but their biases are position, verbosity and self-preference, while ours are about physical state.

Failure benchmarks are the natural place to measure this. However, many of the existing failure benchmarks fix a distribution instead of sampling across one. Episodes are collected in a single campaign, which holds constant the robot and gripper, the camera placement, the scene, the task family, and the procedure that assigned the labels. The three benchmarks shipped by Guardian---RLBench-Fail, BridgeDataV2-Fail, and UR5-Fail \citep{guardian2025}---as well as the
individual benchmarks of REFLECT, RoboFAC, ViFailback, AHA, I-FailSense and RoboReward
\citep{reflect2023,robofac2025,vifailback2025,aha2024,failsense2025,roboreward2026} are each
built this way. ARMOR reports on warehouse rollouts that are not publicly available
\citep{armor2026}. Therefore, a detector scored on one of those benchmarks is scored on one robot performing one
family of tasks under one definition of failure. Section~\ref{sec:results} reports how far
apart a single detector's per-source scores fall once those variables are allowed to move.

Most of these benchmarks also construct their failures: a perturbed demonstration
\citep{aha2024}, an automatically generated error \citep{guardian2025}, a staged mistake
\citep{botfails2026}, or an untouched success relabelled against a different instruction. The
label then follows from the construction procedure, so the failure is visible wherever that
procedure leaves a trace, and a detector can score well by recognizing the trace. These benchmarks also mix execution errors with
planning errors. Guardian and RoboBench keep the two apart
\citep{guardian2025,robobench2025}, and reported numbers usually pool them, which averages a
nearly solved task with an unsolved one.

Episodes of unstaged failures exist and are available, yet nobody has used them for benchmarking. RH20T rates the quality of
roughly 110k teleoperated trajectories \citep{rh20t2024}, and REASSEMBLE labels every
segment of contact-rich assembly, 516 of them real failures \citep{reassemble2025}.
Neither was built to test failure detection, which is why we draw on them, and both turn
out to be among the hardest slices we measure.

The field, meanwhile, builds past the binary label rather than under it: an explanation of
the failure \citep{reflect2023,aha2024,armor2026}, a category and a correction
\citep{guardian2025,robofac2025,vifailback2025,failsafe2025,cfrecovery2026}, a subgoal or
dense progress score \citep{stepeval2025,prmjudge2026,progressreward2026}, a grade on more
than two levels \citep{evalactions2026,roboeval2025,armnetbench2026}. These are the right
ambitions, but every one of those outputs contains the binary label, and none is validated
across sources. PRIMO R1 is the one cross-source result
we are aware of, transferring zero-shot to RoboFail at 67\% after training for process reasoning \citep{primo2026}.

A second family of detectors works from the policy rather than from a
finished recording. SAFE reads a VLA's hidden states and predicts one
failure score that transfers across tasks \citep{safe2025}, and VLA-FAIL adds a
last-layer Mahalanobis distance to a consistency check over successive action chunks
\citep{vlafail2026}. FIPER scores out-of-distribution behaviour in the policy's
embedding space together with an action-chunk entropy \citep{fiper2025}, and FAIL-Detect
distils the policy's inputs and outputs into scalar signals and treats the problem as
sequential out-of-distribution detection \citep{faildetect2025}. Foresight monitors the
latents of an action-conditioned world model \citep{foresight2026}, and GUARD ablates
vision-language KV-cache entries to score how far an action is grounded in what the
model attended to \citep{guard2026}. Sentinel is the hybrid: a statistical consistency
check on the policy's actions for erratic failures, and a VLM reading the observations
for failures of progress \citep{sentinel2024}.

Most calibrate on successful rollouts alone and
set the threshold by conformal prediction, so they need no failure data
\citep{faildetect2025,fiper2025,safe2025}. And accuracy is traded against detection
time, because the aim is to stop or retry a rollout while it is still running. The open
problem is calibration under shift: hidden-state probes lose accuracy under clutter,
lighting, novel objects and reworded instructions, and SAFECAST recovers part of it by
training the probe on contrast-set perturbations \citep{safecast2026}, while Foresight
makes the threshold itself adaptive \citep{foresight2026}.

The alternative to an automated judge is a human one, and the recent real-robot
evaluations show what that costs. RoboArena runs double-blind pairwise comparisons over
600 episodes and seven policies across seven institutions \citep{roboarena2025},
ManipulationNet has a central committee verify every submitted performance
\citep{manipnet2026}, and ArmnetBench has an on-site operator score each rollout as
successful, suboptimal or failure \citep{armnetbench2026}. Human evaluation is also
underpowered as usually practised, with policies typically compared on 20 to 30
real-world trials, too few to separate them with statistical confidence
\citep{suresim2025}.

Evaluations of these detectors exist, and each one is confined to a single source. Guardian,
RoboFAC and ViFailback each score a panel of models on the benchmark released in the same
paper \citep{guardian2025,robofac2025,vifailback2025}, and PRIMO~R1 reports one transfer pair
\citep{primo2026}. No published evaluation scores a detector across independently collected
sources under one protocol, so no published number distinguishes a detector's accuracy from
the source it was measured on. FailBench scores 13 detectors on 14 independently collected sources under one protocol.

\section{FailBench}
\label{sec:benchmark}

FailBench is a collection of curated samples from validated public sources. Some are
established failure detection benchmarks. Others were built for a different purpose entirely
and were never meant to test failure detection: armnetbench and roboarena evaluate policies,
robometer and roboreward score reward models, and rh20t and reassemble are ordinary data
collections. We use each of these
under the labels its own recorders assigned. Every sample in FailBench is one robot attempt, carrying the instruction the robot was given, the visual input the model is shown, and one binary label stating whether
the attempt succeeded.

We screened roughly 30 candidate corpora against three rules and kept \NumSources{},
\NumRealSources{} real and \NumSimSources{} simulated. A source had to record a robot
carrying a manipulation task to completion. It had to supply an outcome label, so that no
label in FailBench is our own judgment of a video. And it had to cover a task family, a robot
or a failure type the set did not already have.

How a failure came about changes what it looks like on video, so we group the sources by three
routes.  \textbf{Organic}, where the failure happened on its own during the attempt:
real policy rollouts in ur5fail (3D-LOTUS++ on a UR5), phail (openpi on a Franka), roboarena
(various policies on DROID) and armnetbench ($\pi_{0.5}$ on SO-101 cells), the same thing
in simulation in simplerenv \citep{simplerenv2024} and robometersim \citep{robometer2026}
where the simulator judged the outcome, and ordinary human teleoperation in rh20t and
reassemble. \textbf{Planned}, where a person drove the arm and made the mistake deliberately:
botfails and robofac, and reflect, whose episodes run on a real arm but follow a script that
fixes the outcome in advance. \textbf{Synthetic}, where the failure was made after the
recording: bdv2fail starts from successful BridgeData~V2 teleoperation and edits the
trajectory, perturbing or reversing it so the arm ends up acting on the wrong object or
leaving it in the wrong place, and roboreward derives its negatives from quality scores on
real clips. One source stays outside the three: robometer documents no verification of how
its labels were assigned, so its origin is marked \emph{undocumented}.

The \NumSources{} sources span single-arm tabletop pick, place and push; kitchen and
household chores; contact-rich assembly on a task board; insertions, cable clipping and tool
handling on low-cost arms; and bin-to-bin picking. Two are multi-robot collections in their
own right, over seven rigs and over five rigs at three universities, and armnetbench
contributes bimanual cells alongside its single-arm ones, so no single robot or laboratory
dominates. Most slices are episode video from one fixed outside camera; two ship three
synchronized views including a wrist camera, two ship an outside and a wrist view, and two
release only a start and an end frame, which is all their authors published.
Table~\ref{tab:sources} in the Appendix~\ref{app:sources} lists every source with its origin and its media.

\begin{figure}[t]
\begin{center}
\includegraphics[width=\linewidth]{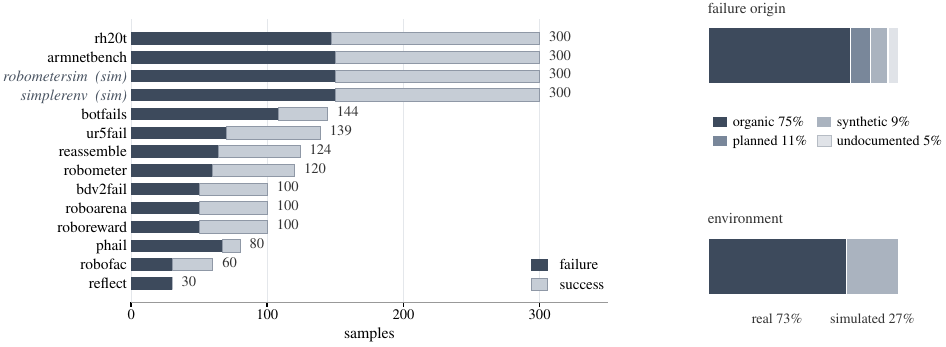}
\end{center}
\caption{FailBench structure. Three
quarters of the benchmark are organic failures. Table~\ref{tab:subsets}
gives what each source had available before sampling.}
\label{fig:composition}
\end{figure}

Some of these sources were collected for purposes other than detector benchmarking. They
therefore contain information that is redundant for a binary outcome question. We subsample
every source rather than taking it whole, while keeping the diversity of scenes, tasks, robots,
and failure types in mind. We draw failures and successes to roughly 50/50, though several subsets keep their
native imbalance because there is not enough data to recover it, and within a source we draw
per failure type so a subset covers the ways that source breaks rather than
over-representing the most common one. Weighted by samples, \FailPct{} of the benchmark is
failures. Real failures are the ones a deployed judge will meet and the expensive ones to
obtain, so the benchmark is built to be majority-organic while keeping enough planned and
synthetic ones to measure whether failures somebody arranged behave differently at all. The
result is \NumSamples{} samples, \NumFailures{} failures and \NumSuccesses{} successes, spread over the
\NumSources{} sources as Figure~\ref{fig:composition} shows.
Appendix~\ref{app:sources} describes each source individually and gives the ledger of what
each one had available against what we drew, in Table~\ref{tab:subsets}.

We randomly select around 25\% of the benchmark for manual inspection. Whenever we observe a
sample with a wrong or ambiguous label, we replace it with a different sample from the same source
and distribution.

\section{Experiments}
\label{sec:experiments}

\subsection{Experimental setup}
\label{sec:protocol}

We evaluate \NumModels{} detectors in three groups. Six are general-purpose vision-language models of
different sizes with no robotics training; they include open- and closed-weight models, most of
which are reasoning models.
Two are trained for embodied robot work but not for failure detection, which separates robotics
training from detection training. Five are purpose-built failure detectors.

Whenever available, we use the official prompt provided by the authors and their recommended
settings for the decoding strategy and thinking budget. All experiments where the same model performance with different pipelines is compared are run with the most possible deterministic setting.   Models that support video input are
supplied directly with the video clip; otherwise, 32 still frames
spaced evenly across the clip are used.

We report balanced accuracy. The slices are not equally balanced, and plain accuracy would reward
a model for guessing whichever answer is more common in a given slice.

\subsection{Results}
\label{sec:results}

Each detector answers one binary question per sample: shown the record of a completed attempt
and the instruction, did it succeed? Every cell of Table~\ref{tab:results} is balanced accuracy, which puts chance at 0.50 regardless of the split. We report two aggregate scores per model. The macro average is the arithmetic mean of the subset scores, so every subset is weighted equally, independently of its sample size. We also report the micro average, which pools every sample the model answered into a single balanced accuracy. Unless specified otherwise, we compare the models using the macro average. reflect contains only failures and therefore has no balanced accuracy of its own; its samples enter the micro average, and it is excluded from the macro average. The panel is six general-purpose models, Gemini 3 Flash \citep{gemini2026}, Gemma-4-31B-it \citep{gemma2026}, GPT-4o \citep{gpt4o2024} and three sizes of Qwen3-VL-Thinking \citep{qwen3vl2026}; two models trained for embodied work, Hy-Embodied-VLM-1.0 and HY-Embodied-0.5 \citep{hyembodied2026}; and five purpose-built detectors, Guardian \citep{guardian2025}, RoboReward-8B \citep{roboreward2026}, ViFailback-8B \citep{vifailback2025}, RoboFAC-7B \citep{robofac2025} and FailSense-Calvin-3B \citep{failsense2025}.

\begin{table}[H]
\caption{\emph{Micro overall} pools all answered samples into one balanced accuracy and
therefore includes the failure-only reflect subset, whereas \emph{macro overall} averages
the 13 two-class subset scores.
\emph{Real} and \emph{Sim} are macro averages over the eleven real and two simulated
two-class subsets, respectively.}
\label{tab:results}
\begin{center}
\scriptsize
\setlength{\tabcolsep}{4pt}
\resizebox{\linewidth}{!}{%
\begin{tabular}{lrrrr}
\toprule
Model & Micro overall & Macro overall & Macro Real & Macro Sim \\
\midrule
Gemini 3 Flash       & 0.74 & \textbf{0.77} & 0.76 & 0.77 \\
Gemma-4-31B-it       & \textbf{0.76} & 0.75 & 0.75 & 0.76 \\
Qwen3-VL-8B-Thinking & 0.69 & 0.69 & 0.68 & 0.75 \\
GPT-4o               & 0.67 & 0.69 & 0.69 & 0.68 \\
Qwen3-VL-4B-Thinking & 0.65 & 0.65 & 0.64 & 0.71 \\
Guardian (thinking)  & 0.61 & 0.63 & 0.62 & 0.64 \\
RoboReward-8B        & 0.59 & 0.62 & 0.62 & 0.61 \\
Hy-Embodied-VLM-1.0  & 0.62 & 0.61 & 0.61 & 0.64 \\
ViFailback-8B        & 0.56 & 0.59 & 0.60 & 0.54 \\
HY-Embodied-0.5      & 0.53 & 0.54 & 0.54 & 0.54 \\
Qwen3-VL-2B-Thinking & 0.53 & 0.53 & 0.52 & 0.55 \\
RoboFAC-7B           & 0.52 & 0.51 & 0.51 & 0.51 \\
FailSense-Calvin-3B  & 0.50 & 0.50 & 0.50 & 0.53 \\
\bottomrule
\end{tabular}}
\end{center}
\end{table}

Table~\ref{tab:results} shows that the best detector among the tested models is \BestModel{},
whose performance reaches \BestScore{}, corresponding to roughly one incorrect judgment in four.
The second-best detector is Gemma-4-31B-it, which scores within a point of the leader (and
outperforms it on the micro average). Every purpose-built detector scores below every
general-purpose model except the smallest, showing that robotics-specific training does not account
for the gap. The two models trained for embodied robot work sit in the same band. Two of the
specialists answer failure almost regardless of what they are shown, which earns one of them a
perfect score on the failure-only slice and leaves both at chance wherever the classes are balanced;
Section~\ref{sec:specialists} compares these models with the base models from which they were
fine-tuned.

Among all real subsets, the worst average performance is observed on reassemble (0.52), where even
the best detector scores only 0.60. The dataset behind this subset covers teleoperated episodes of
contact-rich assembly tasks using the NIST Task Board. We hypothesize that accurate
failure/success detection may be difficult either because of the board-specific knowledge
required or because of the high-precision manipulation of small objects. Among the simulated
subsets, simplerenv has a significantly different distribution of scores from the other subsets. For
example, it is the only subset where Qwen3-VL-8B-Thinking outperforms the top two models.

Simulation is worth separating out, because it is where many detectors are trained and the panel reports it on its own. The real and the simulated halves rank the detectors almost identically, at Spearman $\rho = \SimRealRho{}$, so the simulated slices are not asking a different question; they ask the same one about \SimMinusReal{} points more easily on average. Furthermore, the real environment exhibits larger performance dispersion ($var \approx 0.0135$) compared to the simulation ($\approx 0.0089$). This suggests that simulated environments mask the performance gaps and a judge chosen on them may not perform equally when deployed in a real environment.

\begin{table*}[t]
\caption{Every cell is balanced accuracy,
except reflect ($\dagger$), which contains only failures and is therefore scored by failure
recall.}
\label{tab:results-subsets}
\begin{center}
\scriptsize
\setlength{\tabcolsep}{3pt}
\resizebox{\textwidth}{!}{%
\begin{tabular}{l *{11}{r} !{\vrule width 1.2pt} r !{\vrule width 0.6pt} *{2}{r}}
\toprule
& \multicolumn{12}{c!{\vrule width 0.6pt}}{FailBench-real}
& \multicolumn{2}{c}{FailBench-sim} \\
\cmidrule(lr){2-13} \cmidrule(lr){14-15}
Model & rh20t & armnetbench & botfails & ur5fail & reassemble & robometer & bdv2fail
& roboarena & roboreward & phail & robofac & reflect$^\dagger$ & robometersim & simplerenv \\
\midrule
Gemini 3 Flash       & 0.60 & 0.71 & 0.67 & 0.84 & 0.60 & 0.83 & 0.79 & 0.72 & 0.86 & 0.89 & 0.90 & 0.40 & 0.82 & 0.72 \\
Gemma-4-31B-it       & 0.71 & 0.78 & 0.75 & 0.79 & 0.58 & 0.78 & 0.71 & 0.70 & 0.82 & 0.67 & 0.95 & 0.73 & 0.83 & 0.69 \\
Qwen3-VL-8B-Thinking & 0.62 & 0.68 & 0.61 & 0.62 & 0.49 & 0.73 & 0.63 & 0.71 & 0.72 & 0.77 & 0.90 & 0.57 & 0.71 & 0.79 \\
GPT-4o               & 0.63 & 0.69 & 0.63 & 0.76 & 0.54 & 0.75 & 0.57 & 0.73 & 0.74 & 0.67 & 0.85 & 0.53 & 0.72 & 0.63 \\
Qwen3-VL-4B-Thinking & 0.61 & 0.56 & 0.68 & 0.66 & 0.53 & 0.63 & 0.55 & 0.72 & 0.67 & 0.63 & 0.83 & 0.33 & 0.70 & 0.72 \\
Guardian (thinking)  & 0.51 & 0.57 & 0.59 & 0.74 & 0.53 & 0.70 & 0.85 & 0.57 & 0.61 & 0.50 & 0.70 & 0.30 & 0.69 & 0.59 \\
RoboReward-8B        & 0.63 & 0.66 & 0.56 & 0.51 & 0.50 & 0.53 & 0.52 & 0.69 & 0.71 & 0.70 & 0.82 & 0.00 & 0.48 & 0.73 \\
Hy-Embodied-VLM-1.0  & 0.55 & 0.54 & 0.67 & 0.75 & 0.49 & 0.61 & 0.70 & 0.55 & 0.71 & 0.58 & 0.58 & 0.20 & 0.68 & 0.59 \\
ViFailback-8B        & 0.58 & 0.62 & 0.56 & 0.55 & 0.50 & 0.53 & 0.50 & 0.53 & 0.67 & 0.69 & 0.83 & 0.67 & 0.50 & 0.57 \\
HY-Embodied-0.5      & 0.50 & 0.50 & 0.54 & 0.70 & 0.50 & 0.59 & 0.52 & 0.51 & 0.56 & 0.43 & 0.57 & 0.07 & 0.53 & 0.55 \\
Qwen3-VL-2B-Thinking & 0.51 & 0.52 & 0.58 & 0.56 & 0.51 & 0.53 & 0.46 & 0.54 & 0.55 & 0.50 & 0.50 & 0.23 & 0.53 & 0.58 \\
RoboFAC-7B           & 0.50 & 0.55 & 0.50 & 0.51 & 0.50 & 0.52 & 0.51 & 0.50 & 0.54 & 0.50 & 0.53 & 1.00 & 0.52 & 0.50 \\
FailSense-Calvin-3B  & 0.44 & 0.46 & 0.57 & 0.50 & 0.49 & 0.47 & 0.64 & 0.54 & 0.48 & 0.46 & 0.43 & 0.73 & 0.54 & 0.52 \\
\midrule
Slice mean           & 0.57 & 0.60 & 0.61 & 0.66 & 0.52 & 0.63 & 0.61 & 0.62 & 0.66 & 0.61 & 0.72 & 0.44 & 0.63 & 0.63 \\
\bottomrule
\end{tabular}}
\end{center}
\end{table*}

We dive deeper into diagnosing the benchmark scores and report insights from both qualitative analysis as well as quantitative insights about detector agreement, score validity and fine-tuning impact on specialized detectors in Appendix ~\ref{sec:diagnosis}.

\subsection{Model agreement}
\label{sec:agreement}

The detectors mostly do not fail on the same samples. We measure overlap as intersection over
union (IoU) of error sets, so 1 means identical errors and 0 disjoint ones. The average IoU across all
78 pairs is 0.28 (Table~\ref{tab:overlap}). The strongest overlap, between RoboFAC-7B and
ViFailback-8B at 0.82, comes from the two detectors that answer failure almost regardless of input,
so they are wrong on nearly the same
successes. That is shared bias, not shared perception. Excluding these two detectors, there are 59 samples that are incorrectly classified by the rest of the eleven.

\begin{table}[H]
\caption{Intersection over union of error sets for every pair of detectors. The number in each row label is that
model's column index. The mean over all 78 pairs is 0.28; the maximum, in bold, is the pair
that answers failure almost regardless of input.}
\label{tab:overlap}
\begin{center}
\scriptsize
\setlength{\tabcolsep}{2.5pt}
\begin{tabular}{lrrrrrrrrrrrrr}
\toprule
Model & 1 & 2 & 3 & 4 & 5 & 6 & 7 & 8 & 9 & 10 & 11 & 12 & 13 \\
\midrule
1 \; Gemini 3 Flash      & --- & 0.26 & 0.27 & 0.30 & 0.31 & 0.33 & 0.35 & 0.11 & 0.11 & 0.14 & 0.23 & 0.33 & 0.36 \\
2 \; GPT-4o              & 0.26 & --- & 0.29 & 0.34 & 0.32 & 0.25 & 0.27 & 0.25 & 0.28 & 0.25 & 0.31 & 0.25 & 0.27 \\
3 \; Gemma-4-31B-it      & 0.27 & 0.29 & --- & 0.24 & 0.22 & 0.14 & 0.18 & 0.28 & 0.30 & 0.23 & 0.24 & 0.14 & 0.18 \\
4 \; Qwen3-VL-8B         & 0.30 & 0.34 & 0.24 & --- & 0.40 & 0.36 & 0.33 & 0.15 & 0.16 & 0.19 & 0.26 & 0.30 & 0.35 \\
5 \; Qwen3-VL-4B         & 0.31 & 0.32 & 0.22 & 0.40 & --- & 0.46 & 0.43 & 0.13 & 0.14 & 0.22 & 0.29 & 0.42 & 0.42 \\
6 \; Qwen3-VL-2B         & 0.33 & 0.25 & 0.14 & 0.36 & 0.46 & --- & 0.58 & 0.04 & 0.05 & 0.14 & 0.27 & 0.68 & 0.59 \\
7 \; Guardian            & 0.35 & 0.27 & 0.18 & 0.33 & 0.43 & 0.58 & --- & 0.06 & 0.06 & 0.15 & 0.25 & 0.58 & 0.52 \\
8 \; RoboFAC-7B          & 0.11 & 0.25 & 0.28 & 0.15 & 0.13 & 0.04 & 0.06 & --- & \textbf{0.82} & 0.54 & 0.40 & 0.07 & 0.06 \\
9 \; ViFailback-8B       & 0.11 & 0.28 & 0.30 & 0.16 & 0.14 & 0.05 & 0.06 & \textbf{0.82} & --- & 0.59 & 0.37 & 0.08 & 0.07 \\
10 \; RoboReward-8B      & 0.14 & 0.25 & 0.23 & 0.19 & 0.22 & 0.14 & 0.15 & 0.54 & 0.59 & --- & 0.30 & 0.16 & 0.14 \\
11 \; FailSense-3B       & 0.23 & 0.31 & 0.24 & 0.26 & 0.29 & 0.27 & 0.25 & 0.40 & 0.37 & 0.30 & --- & 0.27 & 0.25 \\
12 \; HY-Embodied-0.5    & 0.33 & 0.25 & 0.14 & 0.30 & 0.42 & 0.68 & 0.58 & 0.07 & 0.08 & 0.16 & 0.27 & --- & 0.59 \\
13 \; Hy-Embodied-VLM-1.0 & 0.36 & 0.27 & 0.18 & 0.35 & 0.42 & 0.59 & 0.52 & 0.06 & 0.07 & 0.14 & 0.25 & 0.59 & --- \\
\bottomrule
\end{tabular}
\end{center}
\end{table}

An overlap of 0.28 suggests that the models treat episodes differently, potentially signaling
differences in knowledge or perception and an opportunity to improve performance by combining
them. Moreover, only 134 samples are misclassified by any of the top three detectors, which
motivates us to test whether an ensemble of them improves performance. We construct a majority
vote of the three strongest detectors; however, it scores 0.78 against 0.76 for the best model alone,
and it reaches only 0.55 on contact-rich assembly. The improvement is therefore marginal compared
with the evaluation cost, which approximately triples, all else being equal.

\subsection{Boosting failure detection performance}
\label{sec:crop}
\label{sec:interventions}

Our diagnosis establishes that the detectors do not read physical state off the pixels they are given, and that whether a peg is seated or the jaws are closed occupies a few dozen pixels of a shot framed to capture the whole workspace. We therefore act on the evidence and build a pipeline out of the same models already being evaluated.

The pipeline consists of two steps: localization and detection. First, a localizer sees four frames spanning the episode and the task sentence, and returns one rectangle around the region that settles whether the attempt worked. It is never given the outcome, so the rectangle cannot encode the answer. Fixed cameras are cropped to that box; wrist and other moving cameras are sent whole. The detector then runs exactly as it does everywhere else in this paper, on the same prompt, frame budget and decoding settings. Only the field of view changes.

\begin{table}[H]
\caption{Pipeline performance with Gemini 3 Flash as both localizer and detector.}
\label{tab:crop}
\begin{center}
\footnotesize
\setlength{\tabcolsep}{5pt}
\begin{tabular}{lrrrrrr}
\toprule
Source & $n$ & original & crop & $\Delta$ & fixes & breaks \\
\midrule
armnetbench           &  300 & 0.723 & 0.743 &   +2.0 &  29 &  23 \\
rh20t                 &  300 & 0.655 & 0.701 &   +4.6 &  41 &  27 \\
robometersim          &  300 & 0.857 & 0.837 &   -2.0 &  25 &  31 \\
simplerenv            &  300 & 0.693 & 0.800 &  +10.7 &  58 &  26 \\
botfails              &  144 & 0.667 & 0.694 &   +2.8 &  13 &   9 \\
ur5fail               &  139 & 0.828 & 0.857 &   +2.9 &   8 &   4 \\
reassemble            &  124 & 0.567 & 0.591 &   +2.3 &  12 &   9 \\
robometer             &  120 & 0.850 & 0.900 &   +5.0 &  11 &   5 \\
bdv2fail              &  100 & 0.780 & 0.740 &   -4.0 &   5 &   9 \\
roboarena             &  100 & 0.740 & 0.810 &   +7.0 &  12 &   5 \\
roboreward            &   69 & 0.840 & 0.860 &   +2.0 &   4 &   2 \\
phail                 &   80 & 0.918 & 0.873 &   -4.5 &   3 &   9 \\
robofac               &   60 & 0.933 & 0.950 &   +1.7 &   2 &   1 \\
reflect$^\dagger$     &   30 & 0.567 & 0.500 &   $-6.7$ &   2 &   4 \\
\midrule
\textbf{Macro overall} & & \textbf{0.773} & \textbf{0.797} & \textbf{+2.3} & \textbf{223} & \textbf{160} \\
\textbf{Micro overall} & \textbf{2166}\footnote{31 samples having wrist only camera (which is not used in the crop experiment) were excluded from the analysis} & \textbf{0.757} & \textbf{0.783} & \textbf{+2.6} & \textbf{223} & \textbf{160} \\
\bottomrule
\end{tabular}
\end{center}
\end{table}

We first test the pipeline using Gemini 3 Flash, the highest-scoring detector on the benchmark, as both the localizer and the detector. Table~\ref{tab:crop} gives the comparison against the same model judging full frames. The mean over the thirteen scored subsets rises \CropGeminiGemini{} points, from 0.773 to 0.797, fixing 223 samples and breaking 160 (exact McNemar $p = 0.0015$). The gain is largest where the workspace is wide and the decisive object is small: simplerenv gains 11 points, roboarena 7 and robometer 5. Appendix~\ref{sec:ablations} shows that this improvement is not attributable to either component alone: both localization and detection contribute, and the full-resolution crop performs best among the tested input transformations.

\section{Limitations}
\label{sec:limitations}

The insights gained from this research are subject to a few limitations that simultaneously point toward compelling directions for future study.

\textbf{The diagnostics are narrow.} The failure-type counts and every reasoning trace we read
come from one model, the strongest we can inspect. The mechanism is established on
Gemma-4-31B-it and assumed rather than shown for the rest of the panel.

\textbf{Scope.} FailBench scores execution-time failures only, and it scores them as one
binary outcome per attempt. It says nothing about whether a detector can spot a bad plan,
anticipate a failure before it happens, name which failure occurred, or estimate progress
during execution \citep{progressreward2026}. The label is the thinnest one a recording can
carry, which is what makes it comparable across \NumSources{} sources and also what caps
what a score on it can tell you.

\textbf{Non-visual streams are not used.} Every number comes from visual input and the
instruction, although four sources record force-torque or audio. Vision-only matches how
these detectors are deployed, but it means the benchmark does not measure whether a detector
could do better with the contact signal that the hardest slices already record.

\textbf{Embodiment.} Every source is a parallel-jaw arm working on a tabletop:
UR5, WidowX, Franka, SO-100 and SO-101, the DROID Frankas of roboarena, and the rigs of
rh20t and robometer. Bimanual manipulation appears only in the 98 two-arm armnetbench
rollouts; there is no humanoid, no mobile manipulation and no dexterous hand. Whether the
contact-level gap we measure looks the same on those platforms is untested.

\section{Future work}
\label{sec:future}

Extending beyond tabletop arms to corpora with humanoid and dexterous-hand data is the next step
in expanding the benchmark.

The benchmark can also be extended to richer labels than a binary outcome. Some of our sources already annotate what went wrong, however we only use them for sampling and error analysis. Evaluating failure-type classification directly would test whether a detector knows why an attempt failed rather than only whether it failed. Incorporating additional modalities, particularly force-torque and audio signals already available in several of the source datasets, would also allow us to measure how much of the remaining difficulty comes from the visual observation itself.

Progress estimation is another promising extension, yet requires dense annotations throughout an attempt rather than a single label at the end \citep{progressreward2026}. These directions can be incorporated into the existing benchmark. Extending the scope to planning time failures assumes evaluation against the action plan rather than the recording and executed trajectories. It requires a separate benchmark and hence we consider it to be a fundamentally different yet related research direction.

The contact-level failures also suggest a more direct direction for model development. The localization pipeline improves performance by removing irrelevant visual context, but it does not teach the detector to infer physical state. Future work should therefore investigate models and training objectives explicitly designed to reason about fine-grained spatial and contact relationships, and evaluate whether such models can close the gap on contact-rich assembly rather than only improve performance on failures whose outcome is already visually obvious.

\section{Conclusion}
\label{sec:conclusion}

We introduced FailBench, a cross-source benchmark for evaluating whether VLMs can reliably judge the success of robot manipulation attempts. Across 2,197 attempts from fourteen public sources, the strongest of thirteen tested detectors reaches only 0.77 mean balanced accuracy. Purpose-built failure detectors consistently underperform general-purpose VLMs, and fine-tuning for failure detection can make a model worse outside the data on which it was trained.

We find that detectors perform substantially better when success can be established from coarse object motion and approach chance when it depends on fine-grained physical contact. Our results show that part of the problem lies in the evidence presented to the model. Localizing the outcome-relevant region and cropping the input improves the strongest detector by 2.3 points without retraining, although the improvement varies substantially across sources and does not solve contact-level failures.

These results suggest that robot failure detection remains an open problem even for current general-purpose VLMs. More importantly, they show that evaluating a detector on the source on which it was developed can substantially overstate its reliability. FailBench provides a common cross-source evaluation for this setting and a basis for measuring progress toward failure detectors that can be trusted when their judgments become rewards, evaluation labels, or decisions about what a robot should do next.


\bibliography{refs}
\bibliographystyle{iclr2027_conference}

\appendix
\section{Description of benchmark subset}
\label{app:sources}
This section describes each source, how its labels are mapped to success and failure, and the sampling criterion used to construct its FailBench subset. Table~\ref{tab:sources} summarizes each source's origin and media. Table~\ref{tab:subsets}
summarizes the eligible pools and selected class counts; the paragraphs that follow give the source-specific splits, filters, and sampling rules.

\paragraph{rh20t \citep{rh20t2024}.} RH20T is a ${\sim}110$k-trajectory teleoperation
corpus spanning 147 tasks and seven collection rigs with different robots, grippers, and
camera configurations. The recording operator assigns each trajectory a quality rating: 0
denotes a robot fault, 1 a completed execution that missed the task goal, and 2--9 a success.
We map ratings 0 and 1 to failure and 2--9 to success. Because RH20T provides no task
taxonomy, we group the curated tasks into 15 manipulation-skill categories and sample 20
scenes per category, targeting 5 robot failures, 5 task failures, and 10 successes where the
available labels permit. The evaluation uses an external-camera video and a wrist-camera
video.

\paragraph{armnetbench \citep{armnetbench2026}.} ArmNetBench evaluates seven policies on
single-arm and bimanual SO-101 systems across 12 tabletop tasks, including stacking,
insertion, removal, clipping, and folding. An on-site operator labels each rollout as
successful, suboptimal, or failed; the release also contains 600 demonstrations without
outcome labels. We retain only $\pi_{0.5}$ policy rollouts because this policy has the highest
success rate among those evaluated. We exclude the demonstrations and the ambiguous
suboptimal tier, then sample 150 failures and 150 successes while balancing labels within
each task. Each sample is shown through three synchronized views: overhead, front, and wrist.

\begin{table*}[t]
\caption{The FailBench sources and the visual evidence shown to each detector. \emph{Failure origin} describes how the negative outcome arose; \emph{rollout generation} describes how the robot motion was produced. View counts refer to the inputs used in our evaluation, not every camera released by the source.}
\label{tab:sources}
\begin{center}
\footnotesize
\setlength{\tabcolsep}{4pt}
\begin{tabular}{lllll}
\toprule
Source & Failure origin & Rollout generation & Visual input & Views shown \\
\midrule
rh20t        & organic      & teleoperation       & video            & 2 (external + wrist) \\
armnetbench  & organic      & policy              & video            & 3 (overhead + front + wrist) \\
robometersim & organic      & policy (simulation) & video            & 1 (third-person) \\
simplerenv   & organic      & policy (simulation) & video            & 1 (third-person) \\
botfails     & planned      & teleoperation       & video            & 1 (high external) \\
ur5fail      & organic      & policy              & start/end frames & 3 external (6 frames) \\
reassemble   & organic      & teleoperation       & video            & 3 (2 external + wrist) \\
robometer    & undocumented & policy              & video            & 1 (fixed external) \\
bdv2fail     & synthetic    & teleoperation       & start/end frames & 1 external (2 frames) \\
roboarena    & organic      & policy              & video            & 2 (external + wrist) \\
roboreward   & synthetic    & mixed               & video            & 1 (source-dependent) \\
phail        & organic      & policy              & video            & 2 (external + wrist) \\
robofac      & planned      & teleoperation       & video            & 1 (external) \\
reflect      & planned      & scripted execution  & video            & 1 (external) \\
\bottomrule
\end{tabular}
\end{center}
\end{table*}

\begin{table}[t]
\caption{What each source offered and what FailBench draws from it. \emph{Available} is the
eligible pool after the source's own splits and our loader's filters; \emph{FailBench} is
the frozen subset. On the available side fail and success need not sum to $n$: 154 rh20t
scenes and one botfails episode carry no usable binary label, and roboarena's pool is
counted before the scene matching its rule requires, so its split is not stated.
armnetbench's pool is the $\pi_{0.5}$ rollouts after its suboptimal tier is dropped.
robometersim's eligible pool is not stated by its source, so the available total covers the
other thirteen. Sources are ordered by the size of the slice they contribute.}
\label{tab:subsets}
\begin{center}
\footnotesize
\setlength{\tabcolsep}{4pt}
\begin{tabular}{l rrr rrr}
\toprule
& \multicolumn{3}{c}{Available} & \multicolumn{3}{c}{FailBench} \\
\cmidrule(lr){2-4} \cmidrule(lr){5-7}
Source & $n$ & fail & succ & $n$ & fail & succ \\
\midrule
rh20t        & 12{,}625 & 838 & 11{,}633 & 300 & 147 & 153 \\
armnetbench  & 345      & 174 & 171      & 300 & 150 & 150 \\
robometersim & ---      & --- & ---      & 300 & 150 & 150 \\
simplerenv   & 2{,}710  & 1{,}041 & 1{,}669 & 300 & 150 & 150 \\
botfails     & 145      & 108 & 36       & 144 & 108 & 36  \\
ur5fail      & 140      & 71  & 69       & 139 & 70  & 69  \\
reassemble   & 4{,}551  & 516 & 4{,}035  & 124 & 64  & 60  \\
robometer    & 257      & 111 & 146      & 120 & 60  & 60  \\
bdv2fail     & 1{,}000  & 500 & 500      & 100 & 50  & 50  \\
roboarena    & 10{,}783 & --- & ---      & 100 & 50  & 50  \\
roboreward   & 2{,}831  & 689 & 2{,}142  & 100 & 50  & 50  \\
phail        & 524      & 493 & 31       & 80  & 67  & 13  \\
robofac      & 1{,}204  & 960 & 244      & 60  & 30  & 30  \\
reflect      & 30       & 30  & 0        & 30  & 30  & 0   \\
\midrule
\textbf{total} & \textbf{37{,}145} & \textbf{---} & \textbf{---}
               & \textbf{\NumSamples{}} & \textbf{\NumFailures{}} & \textbf{\NumSuccesses{}} \\
\bottomrule
\end{tabular}
\end{center}
\end{table}

\paragraph{robometersim \citep{robometer2026}.} RoboMeterSim is the simulated counterpart
of RoboMeter. It contains Meta-World and LIBERO-90 policy rollouts labelled by the
environment's success predicate. We sample 150 failures and 150 successes from the released
RBM-EVAL split, with no overlap with any training configuration. Each sample is a
single-view third-person video.

\paragraph{simplerenv \citep{simplerenv2024}.} SimplerEnv evaluates two RT-1 checkpoints
and RT-2-X in simulated Google-robot scenes. Outcomes are defined by the environment's goal
predicate. We sample 100 rollouts per policy and balance success and failure within policy
and task. Each sample is a single-view third-person video.

\paragraph{botfails \citep{botfails2026}.} BotFails contains teleoperated SO-100 episodes
of kitchen-style tasks with deliberately introduced anomalies. The source provides
per-frame anomaly labels; we label an episode as failed if any frame is annotated as
anomalous. We retain all 144 labelled test episodes, yielding 108 failures and 36 successes. Although the source records elevated
and near-table cameras, the evaluation uses the high external view only.

\paragraph{ur5fail \citep{guardian2025}.} UR5FAIL contains real rollouts of a
3D-LOTUS++ policy performing tabletop pick, place, and push tasks, with human-written labels
for the observed failure mode. We use the execution subset. The resulting subset
contains 70 failures and 69 successes. Each sample is represented by start and end frames
from three external views, for six images in total.

\paragraph{reassemble \citep{reassemble2025}.} REASSEMBLE records teleoperated assembly
and disassembly on the NIST Assembly Task Board~1 using a Franka FR3. Its 153 long-horizon
recordings are segmented into approximately 4{,}551 actions with binary outcome labels. We
use action segments from the official test split and sample toward eight failures from each
of 8 annotated types: Align, Pull, Grasp, Twist, Approach, Lift, Release, Push. The
frozen subset contains 64 failures and 60 successes from the same split. Each segment is
shown through 2 external and 1 wrist cameras.

\paragraph{robometer \citep{robometer2026}.} RoboMeter contributes its held-out
out-of-distribution split, which covers five robot rigs at three universities and assigns a
label of success, partial progress, or failure to each clip. We exclude the ambiguous
partial-progress tier, undocumented duplicate wrist-camera copies, and the one rig with a
moving camera. From the resulting 257 binary-labelled clips, we sample 60 matched
failure--success pairs within rig and task. The source does not document human verification
of its labels, so we manually review the eligible pool before freezing the subset. The
evaluation uses one fixed external-camera video.

\paragraph{bdv2fail \citep{guardian2025}.} BDV2FAIL provides episode-level outcome labels
for BridgeData~V2 WidowX teleoperation in household scenes. Its failures are constructed by
perturbing or reversing successful trajectories. We sample 25 failures labelled ``wrong
object manipulated,'' 25 labelled ``wrong object state or placement,'' and 50 successes.
Each sample contains start and end frames from one external camera.

\paragraph{roboarena \citep{roboarena2025}.} RoboArena is a distributed evaluation on the
DROID platform in which participating laboratories run their policies on a shared robot and
the evaluator records success or failure. To avoid cross-source overlap, we exclude videos
that also appear in any other dataset represented in FailBench. From the remaining pool, we
retain sessions containing both outcomes and sample one failure and one success from each of
50 scene-matched sessions. The failures are split evenly between attempts that reached
approximately one quarter and one half of the task. Each sample is evaluated using
synchronized external- and wrist-camera videos.

\paragraph{roboreward \citep{roboreward2026}.} RoboRewardBench assigns quality scores from
1 to 5 to real-robot OXE and RoboArena rollout clips. We map score 5 to success, scores 2 and
3 to failure, and exclude score 4 because it represents ambiguous near-completion. The
frozen subset contains 50 clips scored 5, 25 scored 3, and 25 scored 2. Each sample uses one
source-dependent camera view.

\paragraph{phail \citep{phail2026}.} PhAIL contains openpi policy rollouts on a real Franka
performing bin-to-bin pick and place across four item categories, with labels derived from
the recorded outcome. The inference split contains only 13 successes, all of which are
included; failures are then sampled up to 20 per item category, yielding 67. Of these
failures, 61 are timeouts. Each sample is evaluated using external- and wrist-camera videos.

\paragraph{robofac \citep{robofac2025}.} RoboFAC is a failure-oriented video-question
answering dataset with simulated and real manipulation splits. We use only the real-robot
split and derive the binary label from its failure-detection answer. The subset contains 10
failures from each of three annotated types---grasping error, orientation deviation, and
position deviation---and 30 successes distributed across all six tasks. Each sample is a
single-view external-camera video.

\paragraph{reflect \citep{reflect2023}.} REFLECT uses the complete public real-robot portion
of RoboFail: 30 scripted household-kitchen episodes, all failures, each accompanied by a
written explanation. We include all 30 episodes. Because the subset contains no successes,
we report failure recall and exclude it from balanced-accuracy means. Each sample is a
single-view external-camera video.

\section{FailBench Diagnosis}
\label{sec:diagnosis}

In this section we dive deeper into the benchmark results and analyze them from three perspectives: whether the scores are valid, how the errors
are distributed across the panel, and what the models report in their own reasoning traces. For the latter, we use Gemma-4-31B-it, the strongest open-weight model in the panel for which we have access to its reasoning traces.

\subsection{Score validity}
\label{sec:specialists}

The purpose-built detectors score below their published numbers, so we first check that the
difference comes from the models rather than from our harness. Those papers report plain accuracy on
their own splits and we report balanced accuracy on a balanced subset, so the two should agree once
normalized. Guardian agrees: 0.741 on UR5FAIL against a published 0.77, and 0.850 on BDV2FAIL
against 0.85.

RoboFAC-7B is where the metrics come apart. Its 0.53 here against a published
\RoboFacSelfReport{} looks like a 27-point shortfall but is not one: its split is 79.7\%
failures and ours is balanced. On our balanced set, RoboFAC-7B labels all 30 failures and 28
of the 30 successes as failures. Its failure and success recalls are therefore 1.000 and
0.067. Reweighting these recalls to the published split---960 failures and 244
successes---gives 0.811 plain accuracy, within half a percentage point of the reported 0.806
\citep{robofac2025}. Its published score is close to what answering
failure to everything already earns, and balancing the classes turns it into chance. The
published ViFailback split has a similar imbalance: 445 failures and 55 successes, so an
all-failure answer scores 0.890, only one point below the 0.900 reported for its Qwen3-VL-8B
base model \citep{vifailback2025}. This is a prior check rather than a reproduction: FailBench contains no subset
from ViFailback. Section~\ref{sec:reproducing} gives the reproduction arithmetic and the accompanying text gives the ViFailback calculation.

We also ran the base models behind the specialized ones under the same prompt, decoding, and input recipe and report them in Table~\ref{tab:bases}. The subset-level values behind Table~\ref{tab:bases} are reported in Table~\ref{tab:app-bases}. All fine-tuned specialists score below the base model they were fine-tuned from. Guardian is the closest to a tie, at 0.627 against 0.629. That near tie is driven by its two home subsets, both of which it wins. Drop those two and score both models on the remaining eleven: Guardian reaches 0.596 against 0.639 for InternVL3-8B.

Base and specialist differ only in the fine-tuning, so the harness, the prompt and the model family do not explain these scores. Fine-tuning for failure detection made these models worse at it outside the data they were tuned on.

\begin{table}[t]
\caption{Each purpose-built detector against the general-purpose model it was fine-tuned
from, scored under the same prompt, decoding and input recipe. Means are over all thirteen
two-class subsets; reflect is excluded because it contains no successes.
$\Delta$ is specialist minus base, computed before rounding.}
\label{tab:bases}
\begin{center}
\footnotesize
\setlength{\tabcolsep}{5pt}
\begin{tabular}{llrrr}
\toprule
Specialist & Base model & Specialist & Base & $\Delta$ \\
\midrule
Guardian (thinking) & InternVL3-8B           & 0.627 & 0.629 & $-0.002$ \\
RoboReward-8B       & Qwen3-VL-8B-Instruct   & 0.619 & 0.675 & $-0.057$ \\
ViFailback-8B       & Qwen3-VL-8B-Instruct   & 0.587 & 0.675 & $-0.088$ \\
RoboFAC-7B          & Qwen2.5-VL-7B-Instruct & 0.514 & 0.555 & $-0.040$ \\
FailSense-Calvin-3B & PaliGemma2-3B          & 0.503 & 0.506  & $-0.003$     \\
\bottomrule
\end{tabular}
\end{center}
\end{table}

\subsection{Manual error analysis}
\label{sec:concentrate}
\label{sec:qualitative}

We inspected the samples the detectors get wrong by hand: the recordings, the instructions, the labels the sources assign, and, where a model emits them, its reasoning traces. Most of the inspection used Gemma-4-31B-it, the strongest open model in the panel, because its traces can be read. The inspection followed no fixed protocol and covered a small share of the errors. Nothing in this subsection is a statistically supported claim.

\begin{figure}[t]
\begin{center}
\includegraphics[width=\linewidth]{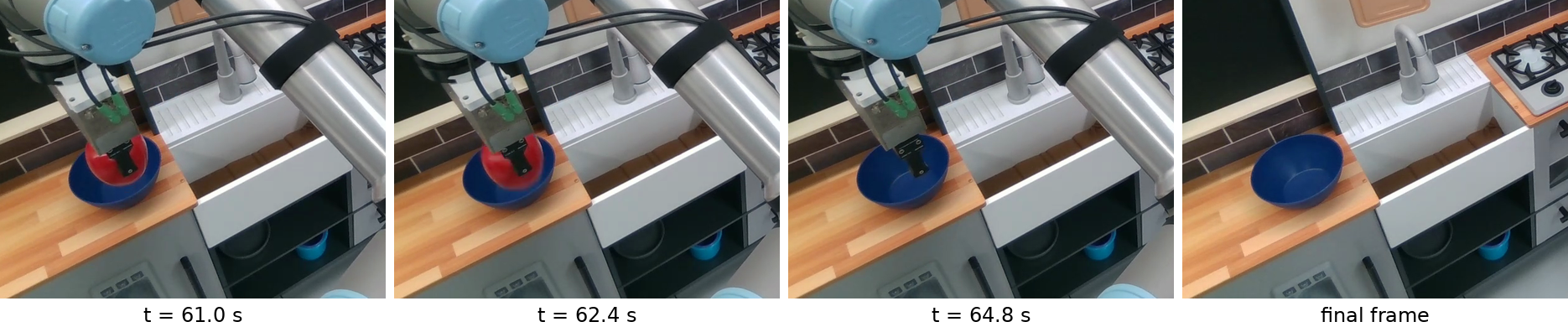}
\end{center}
\caption{Confabulation on REFLECT \texttt{putAppleBowl2}. The apple is dropped from the
gripper at about $t=63$\,s and never reaches the bowl. The bowl is empty in every later frame,
including the last one shown. Gemma-4-31B-it answered success.}
\label{fig:apple}
\end{figure}

Our inspections provide no evidence that a model's accuracy is associated with the robot or the embodiment: the same detector does well and badly on slices recorded on the same class of arm. Nor could we associate it with the task itself, in the sense of pick, place, push, insert, or fold. Where the failure is semantic, the detectors mostly catch it: the robot acts on the wrong object or on an item of the wrong colour, and the outcome is readable from object identity. Where the failures are more physical, the detectors mostly miss them: an object is released in the wrong place, the jaws close on nothing, or a grasp slips. We did not annotate the benchmark according to this distinction, and the sources do not share a failure-type taxonomy that would let us count it, so we state it as a reading of the samples we examined rather than as a measurement. Annotating every sample under one taxonomy and scoring the panel against that split is the extension of the benchmark we intend to make next (Section~\ref{sec:future}).

\begin{figure}[H]
\begin{center}
\includegraphics[width=\linewidth]{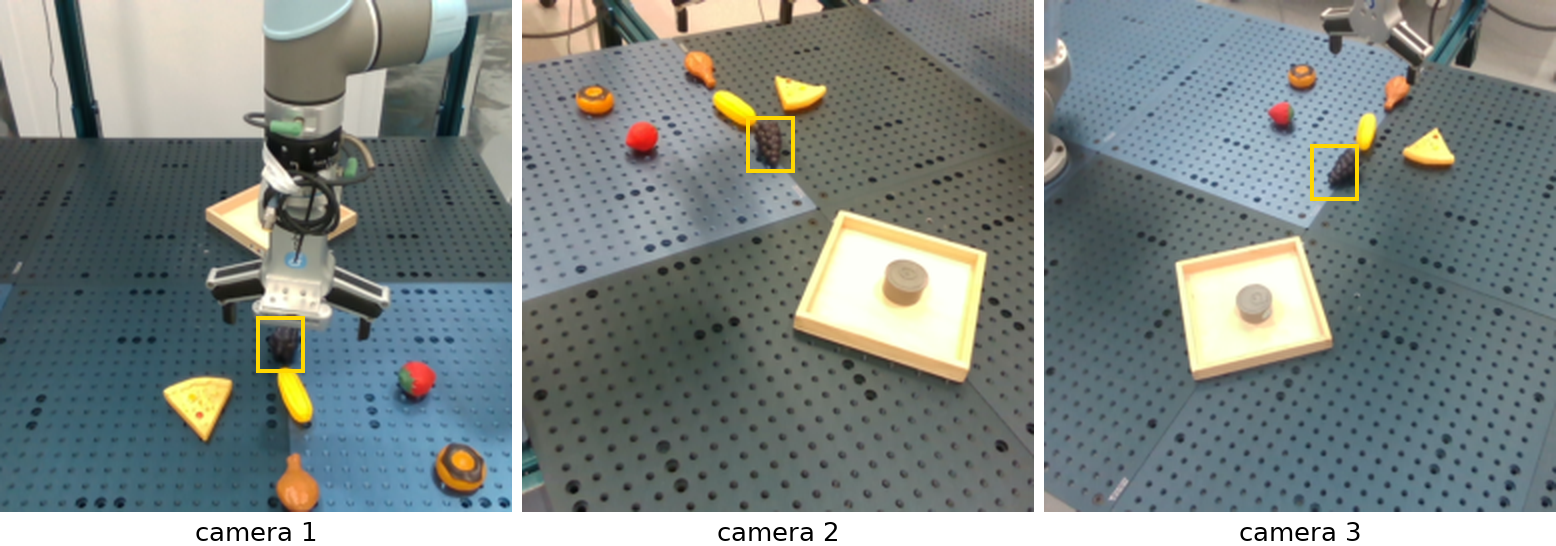}
\end{center}
\caption{A viewpoint failure on ur5fail \texttt{put\_food\_in\_box}. The jaws never closed. Yellow boxes mark the
grapes: from camera 1 they read as held, from cameras 2 and 3 they are still on the table.
All three views go to the model, which answered success.}
\label{fig:viewpoint}
\end{figure}

The traces and the episodes suggest a shared disposition behind the misses. The model frequently performs the perceptual work, names the thing it is unsure about, and then scores success anyway. On a failed stack a trace closes with \emph{``Is there any reason to believe
it failed? No.''}, and on a toy left at the mouth of a drawer it asks \emph{``Is it in the drawer or just at the drawer?''} and never answers. Failure is treated as what must be proven. It also
hallucinates, sometimes reporting observations that the frames do not support. In one
\textsc{reflect} episode, the apple is dropped on the way to the bowl and appears in no later frame
(Figure~\ref{fig:apple}), while the model's own verification step states that the apple is clearly
visible inside the bowl.

The detector seems to score the approach rather than the outcome, calling a bolt inserted after it has been carried to the hole and turned halfway. It checks the commanded motion without checking whether the scene permits the goal at all, passing episodes in which a bowl sits upside down for the whole recording or a knife lies across the opening. And it does not reconcile one view against another: grapes that sit between the jaws from one camera become \emph{``black grapes are now inside the gripper jaws''} (Figure~\ref{fig:viewpoint}) while two other views in the same input show them still on the table. Across these cases the ambiguity comes from the viewing geometry or from an underspecified success criterion, and the model resolves it toward success.

One property of the traces we did count. Gemma-4-31B-it's wrong answers carry longer reasoning than its right ones, 1,868 characters against 1,464 on average, with the medians moving the same way. Spending too little reasoning on the harder samples therefore does not account for these errors.

\section{Boosting Ablations}
\label{sec:ablations}

We run additional experiments to understand whether the improvement from Section \ref{sec:interventions} should be attributed to the localizer or the detector. The results show that the effect belongs to both halves of the pipeline. Keeping Gemini as the localizer and swapping the detector to Gemma-4-31B decreases overall performance by 0.01 point. However, on simplerenv those same boxes are worth $+10.7$ with a Gemini detector and nothing at all with Gemma. Holding the detector fixed to Gemma and swapping the localizer to Gemma decreases the performance by 0.013 more (compared to Gemini localizer and Gemma judge) while still perforing 0.07 better than no localizer Gemma version. Thus, we conclude that neither half carries the result alone: a well-drawn box only pays if the detector can read it, and a strong detector cannot rescue a badly drawn one.

\begin{table*}[t]
\caption{Subset-level attribution across the tested localizer--detector pipelines. Entries
are balanced accuracy. Mean is the macro average over the thirteen two-class subsets.
reflect$^\dagger$ contains only failures, so its entry is failure recall and is excluded
from the mean. None denotes the original full-frame detector without localization.}
\label{tab:crop-attribution-subsets}
\begin{center}
\scriptsize
\setlength{\tabcolsep}{2.2pt}
\resizebox{\textwidth}{!}{%
\begin{tabular}{llllrrrrrrrrrrrrrr}
\toprule
Pipeline & Localizer & Detector & Mean & armnet & rh20t & rbmsim & simpler & botfail & ur5 & reasm. & rbmtr & bdv2 & arena & reward & phail & robofac & reflect$^\dagger$ \\
\midrule
Baseline Gemini & None   & Gemini & 0.773 & 0.723 & 0.655 & 0.857 & 0.693 & 0.667 & 0.828 & 0.567 & 0.850 & 0.780 & 0.740 & 0.840 & 0.918 & 0.933 & 0.567 \\
Gemini--Gemini  & Gemini & Gemini & 0.797 & 0.743 & 0.701 & 0.837 & 0.800 & 0.694 & 0.857 & 0.591 & 0.900 & 0.740 & 0.810 & 0.860 & 0.873 & 0.950 & 0.500 \\
Baseline Gemma  & None   & Gemma  & 0.768 & 0.800 & 0.669 & 0.860 & 0.727 & 0.736 & 0.842 & 0.608 & 0.792 & 0.740 & 0.800 & 0.810 & 0.677 & 0.917 & 0.633 \\
Gemma--Gemma    & Gemma  & Gemma  & 0.775 & 0.803 & 0.752 & 0.880 & 0.733 & 0.718 & 0.727 & 0.625 & 0.858 & 0.650 & 0.790 & 0.850 & 0.754 & 0.933 & 0.800 \\
Gemini--Gemma   & Gemini & Gemma  & 0.787 & 0.807 & 0.709 & 0.857 & 0.727 & 0.764 & 0.849 & 0.634 & 0.867 & 0.740 & 0.780 & 0.830 & 0.754 & 0.917 & 0.800 \\
\bottomrule
\end{tabular}}
\end{center}
\end{table*}

\begin{table*}[t]
\caption{Crop-rendering ablation with Gemini 3 Flash as the localizer and Gemma-4-31B-it as
the detector. \emph{Masked blackout} retains the full frame but blacks out everything outside the
predicted box; \emph{crop} cuts to the box; and \emph{upscaled crop} resizes that cut
to the original frame size. Bold marks the best score in each row.
reflect contains only failures, so its row reports failure recall and is excluded from the subset mean.}
\label{tab:crop-ablation}
\begin{center}
\footnotesize
\setlength{\tabcolsep}{8pt}
\begin{tabular}{lrrrrr}
\toprule
Subset & $n$ & Original & Masked blackout & Crop & Upscaled crop \\
\midrule
armnetbench       & 300 & 0.800 & \textbf{0.830} & 0.807 & 0.823 \\
rh20t             & 300 & 0.669 & 0.702 & 0.709 & \textbf{0.742} \\
robometersim      & 300 & 0.860 & 0.847 & 0.857 & \textbf{0.870} \\
simplerenv        & 300 & 0.727 & \textbf{0.740} & 0.727 & 0.700 \\
botfails          & 144 & 0.736 & 0.759 & \textbf{0.764} & 0.736 \\
ur5fail           & 139 & 0.842 & 0.849 & 0.849 & \textbf{0.850} \\
reassemble        & 124 & 0.608 & \textbf{0.652} & 0.634 & 0.617 \\
robometer         & 120 & 0.792 & 0.842 & \textbf{0.867} & 0.842 \\
bdv2fail          & 100 & \textbf{0.740} & 0.710 & \textbf{0.740} & 0.710 \\
roboarena         & 100 & \textbf{0.800} & \textbf{0.800} & 0.780 & 0.760 \\
roboreward        & 100 & 0.810 & \textbf{0.840} & 0.830 & \textbf{0.840} \\
phail             &  80 & 0.677 & 0.716 & \textbf{0.754} & 0.677 \\
robofac           &  60 & \textbf{0.917} & 0.900 & \textbf{0.917} & 0.900 \\
reflect$^\dagger$ &  30 & 0.633 & 0.700 & \textbf{0.800} & 0.767 \\
\midrule
\textbf{Macro average} & & 0.768 & 0.784 & \textbf{0.787} & 0.774 \\
\bottomrule
\end{tabular}
\end{center}
\end{table*}

That raises the question of whether we could utilize localizer information better by resizing the crop or masking the rest of the frame, which is what we ablated next. We run the ablation using Gemini-Gemma pipeline. Table~\ref{tab:crop-ablation} gives the full subset-level comparison for 4 runs: original (full frame), cropped, cropped and resized, blackout masked (localized part is visible, the rest is blacked out). The original localization pipeline (cutting to the box and leaving it at its own size) reaches 0.787, while keeping the whole frame but darkening everything outside the box reaches 0.784. Cutting to the box and enlarging it back to the original frame size reaches 0.774. Thus, we stick to the native cut, the best of the three as the main configuration.

None of this teaches the model to read contact state. What it does is take away everything the model does not need to look at, which is why darkening the background works almost as well as cutting to it and why enlarging the cut afterwards adds nothing. The gain is concentrated where the deciding region was smallest against its surroundings, and it still depends on whether the judge can use the narrowed view once it is given one.

\section{Reproducing the published-score checks}
\label{sec:reproducing}

Table~\ref{tab:app-bases} expands the averages in Table~\ref{tab:bases}. It uses the same
thirteen two-class subsets; reflect is omitted because it has no successes. In this section, we aim to reproduce published scores on available benchmark subsets. For RoboFAC-7B, the balanced FailBench subset contains 30 failures and 30 successes. The model catches all 30 failures but clears only 2 of the 30 successes, hence
\[
  \operatorname{BA}=\tfrac{1}{2}\left(\tfrac{30}{30}+\tfrac{2}{30}\right)=0.533.
\]
The published real-world split instead contains 960 failures and 244 successes. Holding the two measured recalls fixed and changing only that class prior gives
\[
  \operatorname{Acc}_{\mathrm{published\ prior}}
  =\tfrac{960}{1204}\tfrac{30}{30}+\tfrac{244}{1204}\tfrac{2}{30}
  =0.811,
\]
against the reported 0.806 \citep{robofac2025}. An all-failure answer already obtains $960/1204=0.797$ on that split.

ViFailback provides the same warning about unbalanced published evaluations, but not an
in-domain reproduction. Its test split has 445 failures and 55 successes, so the constant
all-failure rule scores $445/500=0.890$. The paper reports 0.900 for the Qwen3-VL-8B base
\citep{vifailback2025},
only one point above that constant. Because FailBench contains no ViFailback home subset, we
do not project our ViFailback recalls onto this split or claim to reproduce that result.

\begin{table*}[h]
\caption{Balanced accuracy for each specialist and its base model from which it was
fine-tuned. Bold marks a fine-tuned detector whose unrounded score exceeds its base for that row.}
\label{tab:app-bases}
\begin{center}
\scriptsize
\setlength{\tabcolsep}{4pt}
\begin{tabular}{l rr rrr rr rr}
\toprule
& \multicolumn{2}{c}{InternVL3 family}
& \multicolumn{3}{c}{Qwen3-VL family}
& \multicolumn{2}{c}{Qwen2.5-VL family}
& \multicolumn{2}{c}{PaliGemma2-3B family} \\
\cmidrule(lr){2-3} \cmidrule(lr){4-6} \cmidrule(lr){7-8} \cmidrule(lr){9-10}
Subset & Guardian & Base & RoboReward & ViFailback & Base & RoboFAC & Base & FailSense & Base \\
\midrule
rh20t         & 0.506 & 0.596 & \textbf{0.632} & 0.582 & 0.592 & 0.500 & 0.500 & 0.441 & 0.468 \\
armnetbench   & 0.573 & 0.673 & 0.663 & 0.620 & 0.697 & \textbf{0.550} & 0.507 & 0.460 & 0.468 \\
botfails      & 0.588 & 0.676 & \textbf{0.556} & \textbf{0.560} & 0.551 & 0.500 & 0.602 & \textbf{0.569} & 0.437 \\
ur5fail       & \textbf{0.742} & 0.578 & 0.514 & 0.551 & 0.609 & 0.515 & 0.669 & \textbf{0.502} & 0.493 \\
reassemble    & \textbf{0.526} & 0.500 & 0.501 & 0.500 & 0.552 & 0.500 & 0.500 & 0.492 & 0.500 \\
robometer     & 0.700 & 0.717 & 0.525 & 0.525 & 0.683 & 0.517 & 0.525 & 0.467 & 0.508 \\
bdv2fail      & \textbf{0.850} & 0.560 & 0.520 & 0.500 & 0.588 & 0.510 & 0.520 & \textbf{0.640} & 0.580 \\
roboarena     & 0.570 & 0.660 & \textbf{0.690} & 0.530 & 0.680 & 0.500 & 0.510 & \textbf{0.540} & 0.510 \\
roboreward    & \textbf{0.610} & 0.580 & 0.710 & 0.670 & 0.710 & 0.540 & 0.560 & 0.480 & 0.571 \\
phail         & 0.500 & 0.732 & 0.699 & 0.692 & 0.710 & 0.500 & 0.500 & 0.463 & 0.500 \\
robofac       & \textbf{0.700} & 0.500 & 0.817 & 0.833 & 0.917 & 0.533 & 0.683 & 0.433 & 0.517 \\
robometersim  & \textbf{0.687} & 0.653 & 0.483 & 0.500 & 0.673 & 0.523 & 0.523 & \textbf{0.537} & 0.500 \\
simplerenv    & 0.593 & 0.747 & 0.733 & 0.573 & 0.820 & 0.500 & 0.613 & 0.517 & 0.530 \\
\midrule
Mean          & 0.627 & 0.629 & 0.619 & 0.587 & 0.675 & 0.514 & 0.555 & 0.503 & 0.506 \\
\bottomrule
\end{tabular}
\end{center}
\end{table*}

\end{document}